# An Explainable Machine Learning Framework for the Accurate Diagnosis of Ovarian Cancer

Asif Newaz*

Department of Electrical and Electronic Engineering, Islamic University of Technology, eee.asifnewaz@iut-dhaka.edu

Abdullah Taharat

Department of Electrical and Electronic Engineering, Islamic University of Technology, abdullahtaharat@iut-dhaka.edu

Md Sakibul Islam

Department of Electrical and Electronic Engineering, Islamic University of Technology, sakib-ul-islam@iut-dhaka.edu

A.G.M. Fuad Hasan Akanda

Department of Electrical and Electronic Engineering, Islamic University of Technology, fuadhasan@iut-dhaka.edu

Ovarian cancer (OC) is one of the most prevalent types of cancer in women. Early and accurate diagnosis is crucial for the survival of the patients. However, the majority of women are diagnosed in advanced stages due to the lack of effective biomarkers and accurate screening tools. While previous studies sought a common biomarker, our study suggests different biomarkers for the premenopausal and postmenopausal populations. This can provide a new perspective in the search for novel predictors for the effective diagnosis of OC. Genetic algorithm has been utilized to identify the most significant biomarkers. The XGBoost classifier is then trained on the selected features and high ROC-AUC scores of 0.864 and 0.911 have been obtained for the premenopausal and postmenopausal populations, respectively. Lack of explainability is one major limitation of current AI systems. The stochastic nature of the ML algorithms raises concerns about the reliability of the system as it is difficult to interpret the reasons behind the decisions. To increase the trustworthiness and accountability of the diagnostic system as well as to provide transparency and explanations behind the predictions, explainable AI has been incorporated into the ML framework. SHAP is employed to quantify the contributions of the selected biomarkers and determine the most discriminative features. Merging SHAP with the ML models enables clinicians to investigate individual decisions made by the model and gain insights into the factors leading to that prediction. Thus, a hybrid decision support system has been established that can eliminate the bottlenecks caused by the black-box nature of the ML algorithms providing a safe and trustworthy AI tool. The diagnostic accuracy obtained from the proposed system outperforms the existing methods as well as the state-of-the-art ROMA algorithm by a substantial margin which signifies its potential to be an effective tool in the differential diagnosis of OC.

* **Corresponding Author:**
**Address:**   Department of Electrical and Electronic Engineering, Islamic University of Technology, Gazipur, Bangladesh - 1704**.**
**Email:**   eee.asifnewaz@iut-dhaka.edu
**ORCID:**   0000-0002-7524-7564
**Contact:**  +8801880841119

## 1  INTRODUCTION

Cancer is a disease characterized by the abnormal growth and proliferation of cells in the body, caused by mutations in the DNA. These rogue cells can invade healthy tissues and form tumors, leading to serious damage and in some cases, fatal outcomes [1]. The tumors can be either malignant or benign. Malignant tumors can spread into nearby tissues or other parts of the body and form new tumors through a process called metastasis. Benign tumors, on the other hand, do not invade nearby tissues. The most challenging issue for clinical professionals is to differentiate between the two variants of the tumor before it reaches a convoluted stage. The severity of cancer and the efficacy of the treatment highly weigh upon the type of cancer and how early it is diagnosed. Cancer is one of the leading causes of death worldwide and is responsible for millions of deaths each year. According to the World Health Organization (WHO), cancer was held accountable for approximately 10 million deaths worldwide in 2020 [2].

Gynecologic oncologists have to face the challenge of treating a diverse cancer subtype that includes breast cancer, cervical cancer, ovarian cancer, uterine cancer, etc. Ovarian cancer (OC) is rated second in the US in terms of the most common cause of death from gynecological cancer. It also accounts for the lowest survival rate among patients [3]. It is particularly severe as it often goes undetected until it has advanced to later stages. The year 2018 alone had 184,799 fatalities globally due to OC, with 295,414 women receiving the diagnosis [4]. 2.5% of all female malignancies are ovarian cancer. However, 5% of these malignant patients pass away due to poor survival rates. This can be attributed to the absence of early symptoms and the late stage of diagnosis [5]-[6]. Ovarian cancer develops in the ovaries. While the underlying cause of ovarian cancer remains uncertain, several risk factors have been identified that may elevate a woman's likelihood of contracting the illness. Some of the main factors for OC include increasing age, inherited genetic mutations, family history of cancer, and so on. Early detection is the best chance of survival against OC. The five-year survival rate of OC is 92.6 % if diagnosed during stage I, but it drastically drops to 30.3% when diagnosed during stage IV [7]. Therefore, the importance of early detection and research in developing new methods for detecting this cancer in its early stages cannot be overstated.

Several different techniques are implemented for the clinical assessment of OC patients. A pelvic ultrasound test is generally conducted to confirm the existence of cysts during the clinical diagnosis. The resulting image report discloses vital insights into the cancerous potential of the adnexal mass [8]. It performs the classification based on a standard diagnostic score called USG score which incorporates five parameters of the ovarian cyst mass that are extracted from the images - bilaterality, presence of ascites, multilocularity, the existence of solid spots, and signs of intra-abdominal metastasis [9]. Tumor biomarkers provide another way of diagnosing patients with OC. Biomarkers such as carbohydrate antigen 72-4 (CA72-4), human epididymis protein 4 (HE4), and carbohydrate antigen 125 (CA125) are used for the clinical assessment of OC [10]. Acute levels of the serum CA72-4 and CA125 are reported to be found in carcinoma tissues having malicious characteristics. The efficacy of these biomarkers in predicting OC has been investigated, alongside the intervention of new diagnostic indexes involving them. Risk of Ovarian Malignancy Algorithm (ROMA), which incorporates CA125, HE4, and the menopausal status of women was proposed by Moore et al [11]. A similar algorithm, Risk Malignancy Index (RMI) utilizing CA125, the menopausal status of women in conjunction with ultrasound score, was proposed by Jacobs et al. [12]. Ultrasound imaging, tumor biomarkers, and other combined algorithms involving both techniques are generally used in the differential diagnosis of malignant and benign OC masses. Unfortunately, OC often goes undiagnosed until it has progressed to advanced stages due to the lack of specific symptoms in the early stages and the lack of efficient screening methods. This makes OC a stealthy killer and highlights the pressing need for further research to enhance diagnostic and screening methods.

Machine learning (ML) is a powerful tool for analyzing large, complex data and inferring relationships between the feature variables and the target outcome. The use of ML techniques in healthcare has become increasingly popular [13]. It is widely used in medical diagnosis, prognosis, analyzing large electronic health records (EHR), and medical image analysis like CT scans, MRI data, X-ray images, etc. [14]–[16]. ML algorithms have been successfully applied in developing decision-support systems for clinicians to assist in the critical decision-making process [17], [18]. The use of such approaches for the detection of OC can be a promising step forward in the battle against this disease. ML can be extremely useful in detecting cancer in its earliest and most treatable stages, thereby increasing the chances of survival. It can efficiently analyze the patient's data and deduce relationships between patients' characteristics and different test results with the target class (malignant or benign cancer). On that account, a dataset containing the records of 171 OC patients and 178 patients with benign tumors is utilized in this study to develop an ML model that can accurately distinguish malignant tumors from benign ones. The data was originally collected from the Third Affiliated Hospital of Soochow University from 2011 to 2018 [19]. It contains a total of 49 feature variables which include patient demographics, blood routine test results, general chemistry test results, and tumor biomarkers. The goal is to develop a reliable decision-support system with the most appropriate biomarkers to help clinicians predict the presence of OC in patients using ML.

The original data in its raw form needs to be processed to take care of different irregularities present in the data. Once it is done, ML classifiers are then trained on the preprocessed dataset. A k-fold cross-validation approach was employed to test the performance of the classifiers. Compared to traditional classification algorithms like Naïve Bayes (NB), Support Vector Machine (SVM), or Decision Tree (DT), ensemble algorithms like Random Forest (RF) or Extreme Gradient Boosting (XGBoost) are more powerful, accurate, and provide a more generalizable performance. The idea behind ensemble learning is to use a group of weak learners to create a stronger, aggregated one that can provide better performance and generalization. These ensemble learning algorithms are more complex and have been utilized in this study for the prediction of OC.

One of the major issues with applying ML techniques in real-world settings is their black-box nature. Especially in the case of the medical sector, understanding why the model is making a particular prediction and what specific characteristics of the patient are leading to that decision is extremely important. Without this, the model loses its reliability as the clinician has to blindly depend on the decisions provided by the ML model. The lack of such interpretability greatly limits the applicability of ML techniques in the medical field where wrong predictions can have severe consequences. This is a major implementation barrier that prevents clinicians from accepting AI-generated recommendations. To overcome this limitation, we develop an explainable ML framework for the diagnosis of OC. We incorporate the ML algorithm with a popular explainable AI (XAI) framework SHAP (Shapley Additive exPlanations) to provide model-agnostic explanations. XAI is one of the emerging trends in AI as it provides transparent explanations to trust the decisions made by the black-box ML models. SHAP is a unified framework developed to interpret ML model predictions and can be utilized in synergy with any ML algorithm [20]. Interpreting model predictions this way can enable clinicians to better understand the feature variables and their effect on patients' outcomes. It can further elucidate the confidence of the model in predicting a patient's condition and help clinicians assess individual situations explicitly. Such a personalized risk prediction model is a significant step forward for ML in healthcare.

There is also a critical need for identifying potential biomarkers that could lead to the development of novel and more effective predictors for OC diagnosis [21]. This could greatly improve the early diagnosis and survival of OC patients. The dataset at hand contains a total of 49 feature attributes and one response variable. However, not all the features are relevant or related to the target variable. Building an ML model using irrelevant or redundant features gives rise to several problems [22]. The model tends to perform poorly on new data if it is trained on irrelevant features since the information used to develop the model does not represent the target class properly. Also, highly correlated features do not add anything new to the model prediction. The presence of such redundant features further reduces the interpretability of the model. Even with powerful XAI platforms, it becomes difficult for clinicians to analyze the contributions of so many features. The irrelevant features in the data obscure the effect of more critical attributes, ultimately biasing the model toward wrong prediction and interpretation. Feature selection (FS) techniques provide a solution in this regard, allowing the selection of a subset of features that is significant to the target. The use of the most representative subset of features for the training and development of the model ensures an improved and generalized prediction performance. It also enhances the interpretability of the model.

There are different FS algorithms that can be broadly classified into filter and wrapper types [22]. Filter approaches are straightforward variable ranking techniques that utilize correlation or statistical tests. These approaches are computationally light and independent of the underlying classification algorithm. However, they ignore the fact that features can be interrelated, and when combined with other features, they can provide valuable insights into the data. Wrapper methods, on the other hand, are dependent on the underlying classification algorithm and utilize heuristics to obtain sub-optimal subsets of features. They use predictor performance as the objective function to evaluate different subsets of features and obtain the most suitable one. Although these methods are time-consuming, they are often more capable of providing the best representative subset of features. Greedy search-based algorithms like sequential forward selection (SFS), recursive feature elimination (RFE), or population-based metaheuristic optimization algorithms like particle swarm optimization (PSO), and ant colony optimization (ACO) are some of the popular examples of wrapper-type FS methods. In this study, we utilized a powerful evolutionary algorithm to obtain the most suitable subset of features. Genetic algorithm (GA) is an advanced optimization technique for FS that takes inspiration from natural selection [23]. Out of the 49 features, only six features were selected by the GA algorithm. Training the model on only these six features improved the prediction accuracy, while also increasing the explainability of the model.

In summary, the study aims to develop a reliable ML framework that can accurately distinguish malignant ovarian tumors from benign ones and assist clinicians in early diagnosis. The current diagnostic system lacks efficient screening methods and effective biomarkers. Therefore, the genetic algorithm was utilized to identify the most critical risk factors. Training the XGBoost classifier on the selected features, the highest accuracy of 89.6% and 95.8% were achieved on the premenopausal and postmenopausal populations, respectively. One major concern with the use of ML-based diagnostic systems is their black-box nature. Blindly accepting the predictions, especially in safety-critical scenarios like in healthcare, is usually undesirable. To provide transparency, XAI tools have been implemented in this study. This has not been attempted by any of the previous research works on OC diagnosis. To interpret the model predictions and assist clinicians in the decision-making process, SHAP was utilized to provide intuitive explanations behind the contributions of the features to the patients' predicted outcomes. Shapley values can also help in assessing the risk of cancer in patients while the associated visualizations can provide quantitative explanations behind the AI-generated results which can help clinicians make more informed and accurate decisions. This hybrid decision support framework can be more effective, reliable, and robust in the differential diagnosis of OC.

The rest of the manuscript is organized as follows. Related works on AI-assisted OC diagnosis as well as research studies on the identification of OC biomarkers have been discussed in Section – II. In Section – III, the outline of the proposed approach has been discussed. The performance results obtained are presented in Section – IV. The identification of appropriate biomarkers as well as the use of XAI in model interpretation are reviewed in Section - V. We conclude the article in Section VI with a summary of the study.

## 2   RELATED WORKS

Different early screening tools have been developed for the diagnosis of OC using tumor biomarkers. Moore et al. [11] first proposed a dual biomarker algorithm termed ROMA using HE4 and CA125 to classify patients into the high and low-risk categories of OC. Using HE4 and CA125 combinedly is a more accurate predictor of malignancy compared to either alone [24]. In their study, preoperative serum levels of HE4 and CA125 were collected from a total of 531 women who were diagnosed with pelvic mass. The approach was able to successfully calculate the risk of OC in patients, making the ROMA algorithm popular in the risk assessment of OC. RMI is another well-established approach, a classic diagnostic tool used in the preoperative diagnosis of malignancy in women. It utilizes ultrasound test scores along with the serum level of CA125 and the menopausal status. The method was originally proposed by Jacobs et al. [12].

The advantage of ROMA over RMI is the inherent objective nature of the biomarker tests [11]. They are much simpler to conduct. Moore et al. [25] conducted a study on 457 women to compare the performance accuracy of these two approaches. The authors reported that the ROMA score demonstrates a notably higher sensitivity in diagnosing OC in contrast to RMI. In their setting, the ROMA approach achieved a sensitivity of 94.3% compared to RMI which obtained a sensitivity of 84.6% in distinguishing benign tumors from malignant ones.

Anton et al. [26] conducted a prospective study on 128 women to understand and evaluate the contribution of tumor biomarkers CA125 and HE4 in predicting OC. They found HE4 more sensitive to the differential diagnosis of ovarian masses. However, in terms of accuracy, no significant differences were observed between CA125, HE4, ROMA, and RMI. A similar study was conducted by Aslan et al. [27] on 84 women with ovarian masses. They also found HE4 to be the most accurate predictor for diagnosing OC. The ROC-AUC values obtained in the study are - 0.9 for HE4, 0.893 for ROMA, 0.874 for USG (ultrasonography), and 0.794 for CA125.

A meta-analysis was conducted by Wang et al. [28] to assess the diagnostic value of ROMA and the biomarkers HE4 and CA125. The authors systematically analyzed the previous studies and concluded that HE4 is more useful in diagnosing OC in the premenopausal population, while ROMA and CA125 are more appropriate for the postmenopausal population. In a separate study, Terlikowska et al. [29] also attempted to evaluate these biomarkers and the ROMA algorithm based on menopausal status. The conclusions drawn from this study differ from that of Wang et al. [28]. The authors found a high sensitivity of CA125 in detecting OC in the premenopausal population. As for the performance, ROMA was found to provide the highest ROC score. The authors also proposed new cut-off points of ROMA based on their study. The original cut-off values of ROMA for premenopausal and postmenopausal women with high risk are 13.1% and 27.7%, respectively [11]. The proposed new cut-off values from the authors in this study are 14.9% and 33.4% for premenopausal and postmenopausal women, respectively. Different cut-off points were proposed in another study conducted by Zhao et al. [30]. The study was performed on 534 Chinese patients and the proposed cut-off points were 18.47% and 26.48% for premenopausal and postmenopausal women, respectively.

Zhang et al. [31] proposed a new multi-marker model which showed good improvement over the previous approaches. The study was conducted on a population of 149 patients. Along with HE4 and CA125, the authors added two other steroid hormones, progesterone (Prog) and estradiol (E2), in their model. These hormones have been observed in the development of OC [32]. The model was a linear combination of those four markers.

In more recent works, researchers have attempted to incorporate ML in the diagnosis system to improve prediction performance. In that regard, Lu et al. [33] first developed an ML-based prediction system for OC. They conducted their study on a population of 349 patients in China. The authors first utilized the Minimum Redundancy - Maximum Relevance (MRMR) FS method to reduce the number of features from 49 to 10. Of these ten features, only two (HE4 and CEA) were identified as the top features by the DT model. Finally, a logistic regression model was constructed to obtain the final predictions. Their proposed model outperformed the ROMA algorithm with an accuracy rate of 84.7% in comparison to 79.6% from ROMA. In a follow-up study, Ahamad et al. [34] tested the performance of seven different ML algorithms, and the highest accuracy of 88% was obtained using the Light Gradient Boosting Machine (LGBM) classifier.

As can be observed from previous studies, the application of ML as a diagnostic tool for OC has been somewhat sparse. Only a couple of studies have explored the area, but their contributions are quite limited. None of the studies attempted to interpret the model predictions to understand those decisions. Also, there exist contradictory reports in previous studies regarding the biomarkers that are to be used for the diagnosis of OC. Therefore, in this study, we aim to develop an explainable AI framework that can identify the critical risk factors for the accurate detection of OC and provide an ML-based diagnostic tool to assist clinicians in the decision-making process while allowing them to interpret the model predictions for increased reliability and accuracy.

## 3 MATERIALS AND METHODS

### 3.1 Data collection

The data utilized in this study was collected from the Third Affiliated Hospital of Soochow University [19]. It contains the records of a total of 349 OC patients who went through surgical resection from 2011-2018. Among them, 171 patients had malignant tumors, while the remaining 178 patients had benign tumors. None of the patients received any pre-operative radiotherapy or chemotherapy. A statistical summary of the data is listed in Table 1. The data contains a total of 49 feature variables which include blood routine tests, tumor biomarkers, general chemistry tests, and patients' demographics. The target variable is binary representing malignant or benign tumors. The Mann-Whitney U Test was performed to compare the differences between the two groups. The p-values, as well as the mean, median, and range of each attribute, are provided in Table 1.

**Table 1**: Statistical summary of the data

| Feature Name | Acronym | Unit | No. of instances | Mean | Median | Range | p-value |
|---|---|---|---|---|---|---|---|
| Demographics | | | | | | | |
| Age | | | 349 | 45.05 | 45.00 | 15-83 | <0.001 |
| Benign Ovarian Tumors | | | 178 | | | | |
| Ovarian Cancer | | | 171 | | | | |
| Premenopausal | | | 230 | | | | |
| Postmenopausal | | | 119 | | | | |
| Blood Routine Test | | | | | | | |
| Basophil Cell Count | BASO# | 10^9/L | 349 | 0.03 | 0.03 | 0.00-0.12 | 0.312 |
| Basophil Cell Ratio | BASO% | % | 349 | 0.48 | 0.40 | 0.00-1.94 | 0.054 |
| Eosinophil Count | EO# | 10^9/L | 349 | 0.07 | 0.05 | 0.00-0.40 | 0.421 |
| Eosinophil Ratio | EO% | % | 349 | 1.12 | 0.80 | 0.00-7.60 | 0.134 |
| Hematocrit | HCT | L/L | 349 | 0.38 | 0.39 | 0.22-0.57 | 0.033 |
| Hemoglobin | HGB | g/L | 349 | 125.34 | 127.00 | 61.80-189.00 | <0.001 |
| Lymphocyte Count | LYM# | 10^9/L | 349 | 1.56 | 1.50 | 0.35-3.49 | <0.001 |
| Lymphocyte Ratio | LYM% | % | 349 | 26.07 | 26.60 | 3.90-51.60 | <0.001 |

| Feature Name | Acronym | Unit | No. of instances | Mean | Median | Range | p-value |
|---|---|---|---|---|---|---|---|
| Mean Corpuscular Hemoglobin | MCH | Pg | 349 | 28.78 | 29.30 | 17.70-36.80 | <0.001 |
| Mean Corpuscular Volume | MCV | fL | 349 | 88.07 | 89.00 | 61.00-107.90 | 0.414 |
| Mononuclear Cell Count | MONO# | 10^9/L | 349 | 0.36 | 0.32 | 0.07-0.97 | <0.001 |
| Monocyte Ratio | MONO% | % | 349 | 5.58 | 5.43 | 0.30-21.30 | 0.068 |
| Mean Platelet Volume | MPV | fL | 347 | 10.04 | 10.30 | 5.06-14.50 | 0.407 |
| Neutrophil Ratio | NEU | % | 258 | 66.58 | 66.75 | 37.20-92.00 | <0.001 |
| Thrombocytocrit | PCT | L/L | 347 | 0.25 | 0.24 | 0.07-0.69 | <0.001 |
| Platelet Distribution Width | PDW | % | 347 | 14.33 | 13.70 | 8.80-22.80 | 0.002 |
| Platelet Count | PLT | 10^9/L | 349 | 255.43 | 236.00 | 74.00-868.00 | <0.001 |
| Red Blood Cell Count | RBC | 10^12/L | 349 | 4.36 | 4.37 | 2.62-6.74 | 0.031 |
| Red Blood Cell Distribution Width | RDW | % | 349 | 13.55 | 13.10 | 10.92-22.20 | 0.376 |
| General Chemistry | | | | | | | |
| Anion Gap | AG | mmol/L | 348 | 19.32 | 19.85 | 6.20-33.33 | 0.339 |
| Albumin | ALB | g/L | 339 | 41.08 | 42.00 | 22.00-51.50 | <0.001 |
| Alkaline Phosphatase | ALP | U/L | 339 | 77.09 | 71.00 | 26.00-763.00 | <0.001 |
| Alanine Aminotransferase | ALT | U/L | 339 | 18.01 | 15.00 | 4.00-86.00 | 0.353 |
| Aspartate Aminotransferase | AST | U/L | 339 | 19.11 | 17.00 | 7.00-78.00 | <0.001 |
| Blood Urea Nitrogen | BUN | mmol/L | 349 | 4.01 | 3.83 | 1.12-10.19 | 0.449 |
| Calcium | Ca | mmol/L | 349 | 2.39 | 2.47 | 0.92-2.83 | <0.001 |
| Chlorine | CL | mmol/L | 349 | 100.81 | 100.90 | 84.60-109.40 | 0.34 |
| Carbon Dioxide-combining Power | CO2CP | mmol/L | 348 | 24.28 | 24.05 | 16.20-34.30 | 0.329 |
| Creatinine | CREA | µmol/L | 349 | 64.25 | 63.30 | 38.20-114.00 | 0.06 |
| Direct Bilirubin | DBIL | µmol/L | 339 | 3.13 | 2.80 | 0.90-12.10 | <0.001 |
| Gama Glutamyl transferase | GGT | U/L | 339 | 21.31 | 16.00 | 4.00-176.00 | 0.007 |
| Globulin | GLO | g/L | 339 | 30.18 | 30.10 | 14.10-47.60 | 0.001 |
| Glucose | GLU. | mmol/L | 349 | 5.33 | 5.08 | 3.57-12.44 | 0.014 |
| Indirect Bilirubin | IBIL | µmol/L | 339 | 5.96 | 5.40 | 1.00-28.40 | <0.001 |
| Kalium | K | mmol/L | 349 | 4.39 | 4.37 | 3.08-5.40 | 0.908 |
| Magnesium | Mg | mmol/L | 349 | 0.98 | 0.97 | 0.65-1.37 | 0.571 |
| Natrium | Na | mmol/L | 349 | 140.49 | 140.50 | 125.1-150.7 | <0.001 |
| Phosphorus | PHOS | mmol/L | 349 | 1.12 | 1.12 | 0.57-1.75 | 0.641 |
| Total Bilirubin | TBIL | µmol/L | 339 | 9.09 | 8.40 | 2.5-38.3 | <0.001 |
| Total Protein | TP | g/L | 339 | 71.08 | 72.50 | 32.9-86.8 | 0.015 |

| Feature Name | Acronym | Unit | No. of instances | Mean | Median | Range | p-value |
|---|---|---|---|---|---|---|---|
| Uric Acid | UA | µmol/L | 349 | 243.71 | 235.40 | 96.0-632.0 | 0.982 |
| Tumor Marker | | | | | | | |
| Alpha-fetoprotein | AFP | ng/mL | 327 | 11.82 | 2.28 | 0.61-1210.0 | 0.012 |
| Carbohydrate Antigen 125 | CA125 | U/mL | 332 | 350.38 | 44.68 | 3.75-5000.0 | <0.001 |
| Carbohydrate Antigen 19-9 | CA19-9 | U/mL | 325 | 46.73 | 14.20 | 0.6-1000.0 | 0.2 |
| Carbohydrate Antigen 72-4 | CA72-4 | U/mL | 109 | 10.17 | 2.37 | 0.2-158.5 | <0.001 |
| Carcinoembryonic Antigen | CEA | ng/mL | 327 | 3.31 | 1.33 | 0.2-138.8 | 0.103 |
| Human Epididymis Protein 4 | HE4 | pmol/L | 329 | 183.95 | 53.27 | 16.71-3537.6 | <0.001 |

### 3.2 Data preprocessing

The original data was first subjected to certain preprocessing steps to remove irregularities from the data. These include data cleaning, missing data handling, and outlier processing. The ID column was first omitted as it is of no use in the model development stage. The data contained some characters that were removed. As can be observed from Table 1, several feature variables like NEU, MPV, or CA19-9 have missing entries. These entries were imputed using the MICE (Multivariate Imputation by Chained Equation) algorithm. It is quite a popular and sophisticated imputation technique that uses regression to fill in the missing values [35]. It models each feature with missing values as a function of other features and uses that estimate for imputation. Several attributes also contained outliers. Outliers are data points that differ significantly from other observations [36]. They can occur due to some measurement error or natural deviations in the population. The presence of such extreme values in the data can bias the model predictions and affect the interpretability of the model. Figure 1 illustrates the presence of outliers in different attributes in the data using box plots. Such outlier values were truncated to the lower or upper thresholds in the box plot. These thresholds are defined as ± 1.5 times the IQR (Inter-quartile range) from the upper or lower quartiles. After that, each feature variable in the data was standardized using the z-score which can be defined as follows –

$$z - score = \frac{x - \mu}{\sigma} \text{ where } \mu = mean, \text{ and } \sigma = standard\ deviation \tag{1}$$

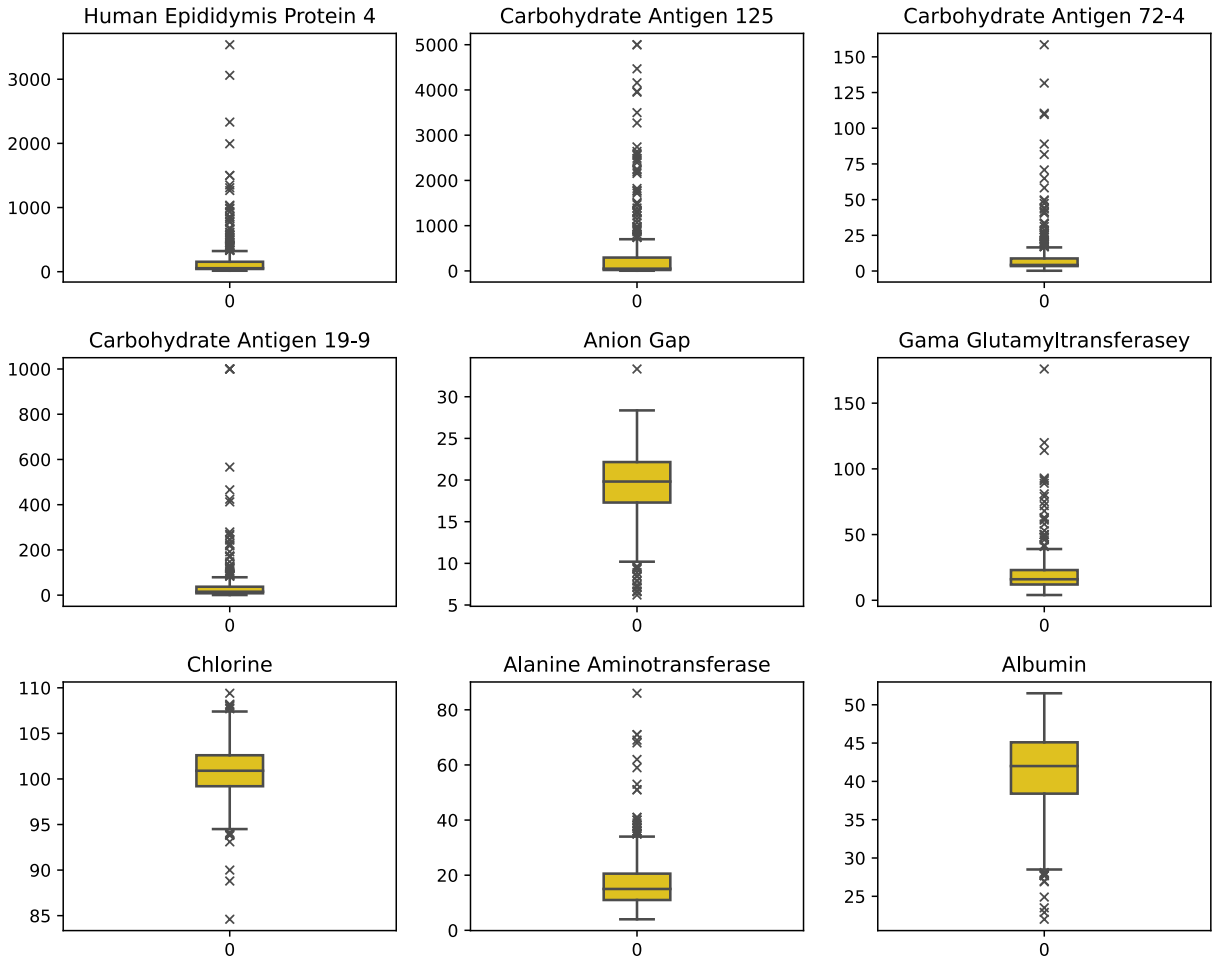

**Figure 1:** Box plot representing the presence of outliers in different features

### 3.3 Proposed methodology

After preprocessing the dataset, it was put into the ML pipeline. First, an FS technique was employed to identify the most significant features in the data. The genetic algorithm was chosen for this purpose. GA is a wrapper approach i.e., it is wrapped around a classification algorithm and iteratively evaluates the model's performance to identify the best features. The XGBoost classifier was selected as the ML algorithm for classification. The model was trained using the selected features from the GA. The menopausal status of the patients has been universally identified as a key risk factor in the occurrence of OC and different cut-off values of the biomarkers are considered for premenopausal and postmenopausal patients [37]. Standard OC risk prediction algorithms like ROMA also consider premenopausal and postmenopausal patients separately. In that regard, two separate ML models were developed for the premenopausal and postmenopausal populations. The FS procedure was also performed separately, and different feature sets were identified for premenopausal and postmenopausal patients. A K-fold cross-validation strategy was utilized to properly assess the performance of the

XGBoost classifier. The value of K was taken as 10. Seven different metrics were considered to thoroughly evaluate the performance of the classifier. SHAP was then used to interpret the model predictions and understand the contribution of each feature. Different visualization techniques were employed for illustration and efficient analysis. For global interpretability, variable importance plots and partial dependence plots were utilized. For local interpretability i.e., to understand the decision made by the model for a particular patient, the SHAP force plot was utilized. The wrong predictions made by the model were also assessed using SHAP. This can help in identifying the reasons behind the wrong predictions. Understanding the factors behind wrong predictions made by the ML model can help in building a more robust decision-support system. Finally, the prediction performance obtained from the proposed approach is compared with the previously reported studies and the ROMA algorithm. The outline of the proposed methodology is illustrated in Figure 2.

### 3.4 Feature selection using genetic algorithm

The use of FS techniques is essential as they allow the selection of features that are appropriate for the ML model while eliminating redundant and irrelevant features. The GA is one of the most popular and advanced optimization techniques for FS. Influenced by the process of evolution, this algorithm utilizes several biologically inspired operators like crossover, mutation, and selection to identify the most significant features in the data [38]. Starting with a population of candidate solutions (subset of features), the model undergoes processes like recombination and mutation to produce new offspring (new subset of features). Each candidate solution has a set of properties called chromosomes which are altered and mutated during the evolution process. While the evolution process starts with a random population i.e., random set of features, at each iteration, a new population is produced called a generation. The population in each generation is evaluated using an objective function. Every individual (feature) in the population is assigned a fitness value based on the objective function. The fitter individuals in the population are stochastically selected using the tournament selection mechanism. These individuals have a higher chance to mate and produce new offspring. To ensure diversity and more coverage of the feature space, the crossover operator is utilized. It performs the XOR operation on the parent chromosomes to produce new offspring that are slightly different than their parents. To further increase the exploratory behavior of the algorithm, the mutation operator is utilized. It inserts random tweaks in the genes of the chromosomes to introduce diversity in the offspring. Finally, the elitist selection mechanism was utilized to propagate the best-fitted individuals to the next generation without any alteration. The new population generated from these processes is again evaluated and passed to the next iteration. Thus, the population evolves over multiple generations, and better-suited individuals (features) are obtained.

Table 2 outlines the parameter values utilized in the development of GA in our study. A grid-search technique was employed to select the most suitable values for the parameters. Here, the number of population refers to the quantity of potential solutions which is represented as individual solutions that exist within each generation of the algorithm's iterative process. This population constitutes the pool from which new solutions are evolved and generated. A larger population size increases exploration through more extensive exploration of the solution space. However, it also results in increased computational demands. The probability of crossover refers to the likelihood that two individuals in the population will exchange genetic information to produce new offspring during the reproduction phase of the algorithm. This probability determines how often crossover (also known as recombination) occurs as a part of the GA operations. It regulates the balance between exploration and exploitation in the optimization process. This balanced probability maintains a crucial equilibrium within exploration and exploitation. Higher values of this probability tend to promote convergence by exploiting promising solutions, while lower values encourage exploration by allowing for more diverse offspring. The probability of mutation represents the likelihood that an individual's genetic material will undergo random changes during the reproduction phase of the algorithm. This plays a vital role in a GA by introducing randomness and diversity into the population. This contributes to exploration by preventing the algorithm from getting stuck in local optima and allowing it to explore different regions of the solution space. Independent probability for each attribute to be exchanged indicates the likelihood that an individual attribute will participate in a crossover operation; independent of other attributes. It ensures that the crossover operation is conducted fairly and uniformly across all attributes within an individual. This type of equitable treatment fosters comprehensive exploration of attribute combinations. Independent probability for each attribute to be mutated indicates the likelihood that each attribute within a solution will undergo random mutations independently during reproduction. It nurtures diversity in the population and helps the GA explore different attribute combinations effectively. Tournament size means the number of individuals randomly selected from the population to compete in a selection tournament. This is done based on their fitness and the fittest among them is chosen for reproduction. It influences the selection pressure and the quality of chosen individuals for reproduction. The tournament size is set to 3. It means during each selection process, three individuals are randomly selected and pitted against each other based on their fitness. The fittest individual among these three is then chosen for reproduction. This setting ensures that selection pressure is moderate, maintaining a balance between favoring better-performing individuals and exploring genetic diversity within the population.

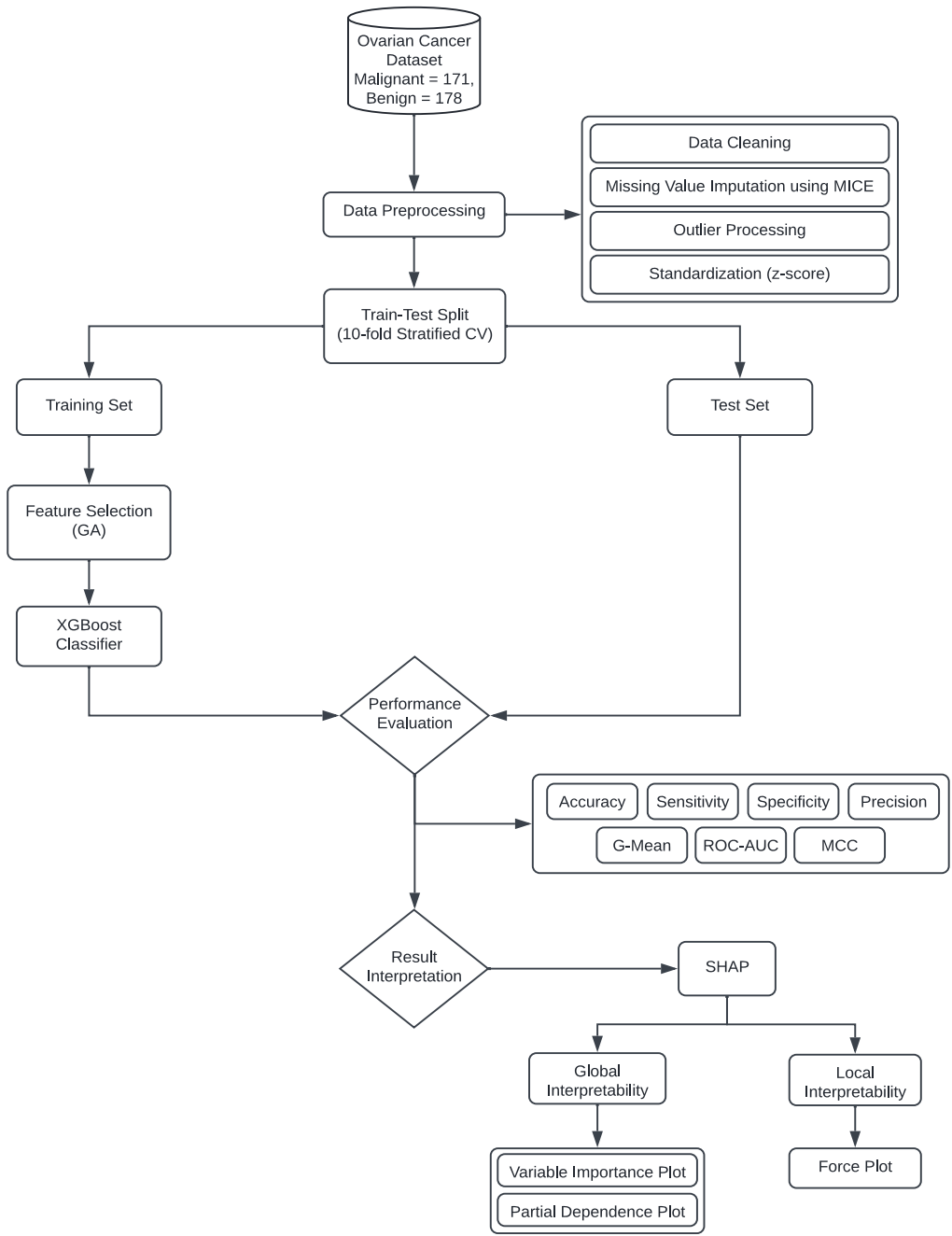

**Figure 2:** Outline of the proposed framework

**Table 2**: GA parameters

| Parameters | Value |
| --- | --- |
| Number of population | 100 |
| Probability of crossover | 0.5 |
| Probability of mutation | 0.2 |
| Independent probability for each attribute to be exchanged | 0.5 |
| Independent probability for each attribute to be mutated | 0.5 |
| Tournament size | 3 |
| Number of generations | 20 |
| Objective function | Accuracy score |

### 3.5 Machine learning algorithm - XGBoost classifier

Ensemble learning refers to the use of multiple classification algorithms as base learners to combinedly produce a stronger predictive model. Base learners (also called weak learners) are usually more susceptible to noise and prone to overfitting. These learners can be strategically combined to produce a more robust ML algorithm. Bagging and boosting are two popular ways of creating an ensemble. In bagging, a group of weak learners is first trained on a random subset of the dataset whose predictions are then aggregated by voting or averaging to obtain the final prediction. Boosting algorithms, on the other hand, seek to improve prediction performance by training a sequence of weak learners, each compensating for the weakness of its predecessors. These models are wildly popular as they often outperform traditional classification algorithms. Adaptive boosting (AdaBoost), or Gradient boosting (GB) are some of the most popular boosting techniques.

XGBoost, which stands for eXtreme Gradient Boosting, is an advanced extension of the GB algorithm, originally proposed by Chen et al. [39]. It has become one of the most popular ML algorithms due to its state-of-the-art performance in many ML tasks [40], [41]. It is a highly sophisticated algorithm designed for parallel processing and is capable of handling very large datasets. An inherent advantage of XGBoost lies in its capability to effectively manage model complexity and counteract the issue of overfitting by employing regularization techniques, ensuring more accurate predictions and enhanced performance. It optimizes computation and memory utilization through parallelization which significantly speeds up training compared to other GB algorithms. It offers a wide range of tunable parameters to control overfitting and increase the generalizability of the model. The versatility and reliability offered by this model have led to the widespread adoption of this technique in many ML applications. In this study, the XGBClassifier from the "xgboost" python package with default parameter settings has been utilized.

### 3.6 Explainable AI with SHAP

A very sophisticated ML algorithm can produce highly accurate predictions, but the black-box nature of the predictions significantly limits its adaptations in real-world applications. Such a lack of transparency is unacceptable in many applications, such as those in the medical industry, where failure might have devastating results. Therefore, in recent years, there has been growing attention to the interpretation of ML models [42]. XAI is an emerging trend in AI which can be defined as a set of processes and methods that allows users to comprehend the output produced by the ML algorithms [43]. This increases the trustworthiness and accountability of the AI system, making it more reliable as a clinical decision support system.

SHAP is one of the popular tools used in XAI and it has been utilized in this study for model interpretation. SHAP is a unified framework based on game theory, originally proposed by Lundberg et al. [20]. It calculates the Shapley values for each sample as well as each attribute in the dataset to provide model-agnostic explanations. Shapley value is a solution concept used in game theory to fairly distribute both gains and costs to several players working in a coalition [44]. By using the Shapley values, SHAP can simultaneously provide both global and local interpretations i.e., the relations of each feature with the target variable and the contributions of the features in predicting each observation, respectively. The globally important features may not always be important in the local context and vice versa. The local interpretations obtained from SHAP can explain why a particular case receives its prediction and which attributes contribute to what degree in making that prediction. This unique feature of SHAP allows for the development of a personalized risk prediction model which can greatly benefit clinicians in predicting the risk more accurately.

The SHAP module in Python offers a variety of visualization techniques for effective interpretation. This includes the partial dependence plot which shows the marginal effect of one or two features on the predicted outcome, the variable dependence plot which shows the positive or negative relationships of the predictor variables with the target variable, the force plot which can be used to interpret individual predictions for different patients, etc. These visualization techniques are utilized in this study to analyze the model predictions.

### 3.7 ROMA Score

ROMA is a dual biomarker algorithm proposed by Moore et al. [11] which is used in the classification of pelvic masses. It is a popular approach in OC malignancy tests which is calculated using the concentrations of the serum biomarkers HE4 and CA125 in conjunction with the menopausal status of the patient. The predictive index of ROMA is determined using the following formulae:

$$\text{Pre-menopausal Predictive Index (PI)} = -12.0 + 2.38*LN(HE4) + 0.0626*LN(CA\ 125) \qquad (2)$$

$$\text{Post-menopausal Predictive Index (PI)} = -8.09 + 1.04*LN(HE4) + 0.732*LN(CA\ 125) \qquad (3)$$

$$\text{The predicted probability of the ROMA score} = \frac{\exp(PI)}{1+\exp(PI)} * 100 \qquad (4)$$

The cut-off value of ROMA score for high risk of OC is 13.1% and 27.7% based on the premenopausal and postmenopausal status, respectively, according to [11]. We calculated the ROMA score for the patients in the dataset accordingly during the experimentation. The scores are then compared with the results from our proposed approach.

## 4 RESULTS

In this study, we propose an ML-based diagnostic system for the prediction of the risk of OC. The XGBoost classifier was chosen for the classification task as it outperforms other traditional classification algorithms. A comparison of the performance among these algorithms in terms of the MCC score is illustrated in Figure 3. The models were trained on the premenopausal and postmenopausal patients separately. From the figure, KNN can be found as the lowest-performing classifier. Ensemble algorithms like RF or AdaBoost classifier performed comparatively better. However, the XGBoost classifier attained the best performance for both groups and was therefore utilized in subsequent processes. The Genetic algorithm was then employed to identify the most representative set of features that optimize the classification accuracy of the XGBoost model. A 10-fold cross-validation strategy was employed to obtain the results. SHAP was then used to assess the model predictions and understand the components that drive the predictions.

### 4.1 Performance of the XGBoost classifier

The performance of the XGBoost classifier, when trained on the original 48 features, is reported in Table 3. A high prediction accuracy of 86.1% and an ROC-AUC score of 0.824 was achieved using this ensemble algorithm for premenopausal patients. For postmenopausal women, the accuracy was even higher 94.9%, and the ROC-AUC score was 0.906. The MCC score is a more reliable and robust metric as it produces a high score only if the prediction obtained good results in all of the four confusion-matrix categories. With the proposed approach, the average MCC score obtained for the population is 76.4%.

### 4.2 Performance after incorporating feature selection strategy

Identifying the most critical risk factors is crucial for effective diagnosis. A large number of features make the decision-support system complicated and necessitates a lot of additional testing that can be quite costly. It also inserts bias into the prediction system and causes overfitting. Therefore, to alleviate the problems, the GA-based FS technique was employed. The model was trained again on the selected features and a good improvement in performance was observed. As can be seen from Table 3, the accuracy score improved to 89.6% and the ROC-AUC score to 0.864 for premenopausal patients. For postmenopausal patients, the sensitivity score reached 98.9% - nearly perfect in accurately predicting the risk of cancer, which is crucial for a reliable medical diagnostic system. The improvement in performance after applying FS is illustrated in Figure 4.

As for the features selected by the GA, different features were found to be more suitable for predicting cancer in premenopausal and postmenopausal women. The selected features from the GA are reported in Table 4. For premenopausal women, six features were identified as the most significant for predicting the risk of OC, while for postmenopausal women, seven features were selected. However, to what extent each of those features contributes towards the prediction of OC is not apparent from the algorithm. That is why the SHAP algorithm is incorporated to obtain a better understanding of the effect of those features on the target variable.

Table 3: Performance measures obtained from the proposed approach

| Approach | Menopause Status | Accuracy | Sensitivity | Specificity | Precision | G-mean | ROC-AUC | MCC |
|---|---|---|---|---|---|---|---|---|
| XGBoost | Premenopausal | 0.861 | 0.712 | 0.935 | 0.861 | 0.808 | 0.824 | 0.686 |
| | Postmenopausal | 0.949 | 0.979 | 0.833 | 0.962 | 0.889 | 0.906 | 0.841 |
| | Total Average | 0.905 | 0.845 | 0.884 | 0.912 | 0.849 | 0.865 | 0.764 |
| XGBoost + GA | Premenopausal | 0.896 | 0.773 | 0.955 | 0.889 | 0.857 | 0.864 | 0.757 |
| | Postmenopausal | 0.958 | 0.989 | 0.833 | 0.963 | 0.893 | 0.911 | 0.868 |
| | Total Average | 0.927 | 0.881 | 0.894 | 0.926 | 0.875 | 0.887 | 0.813 |

Table 4: Features selected by the GA

| Menopausal Status | Features | | | | | | |
|---|---|---|---|---|---|---|---|
| Premenopausal | CA125 | HE4 | CEA | ALB | TP | LYM | |
| Postmenopausal | CA125 | CA199 | RBC | HCT | K | MONO% | Na |

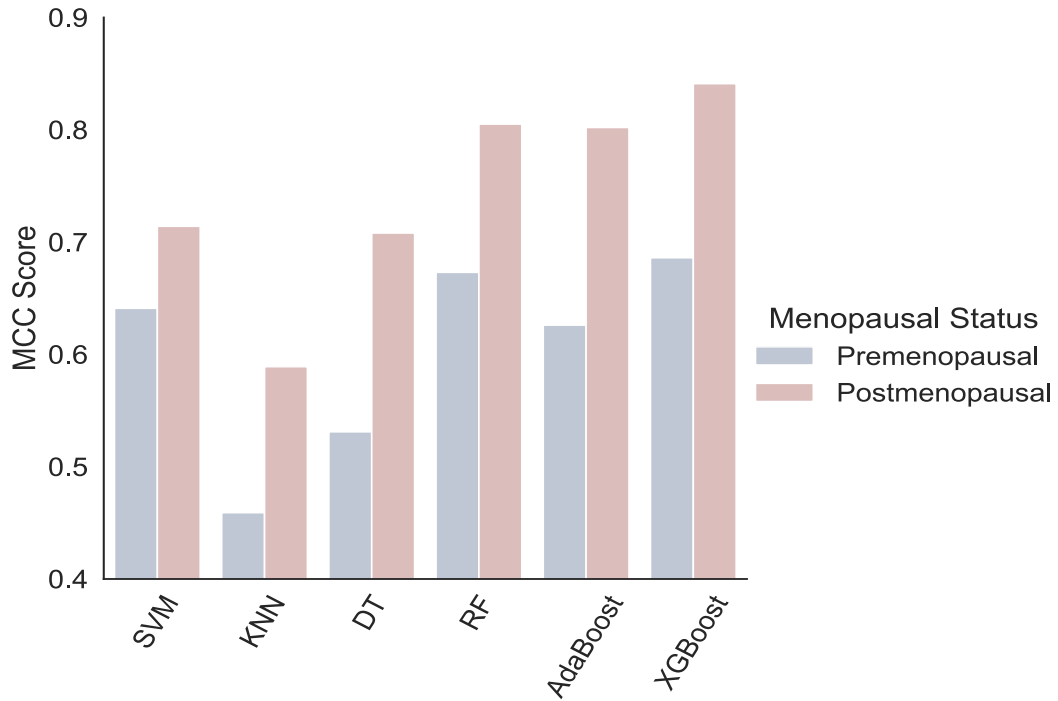

**Figure 3:** Performance comparison of the XGBoost classifier with other classification algorithms

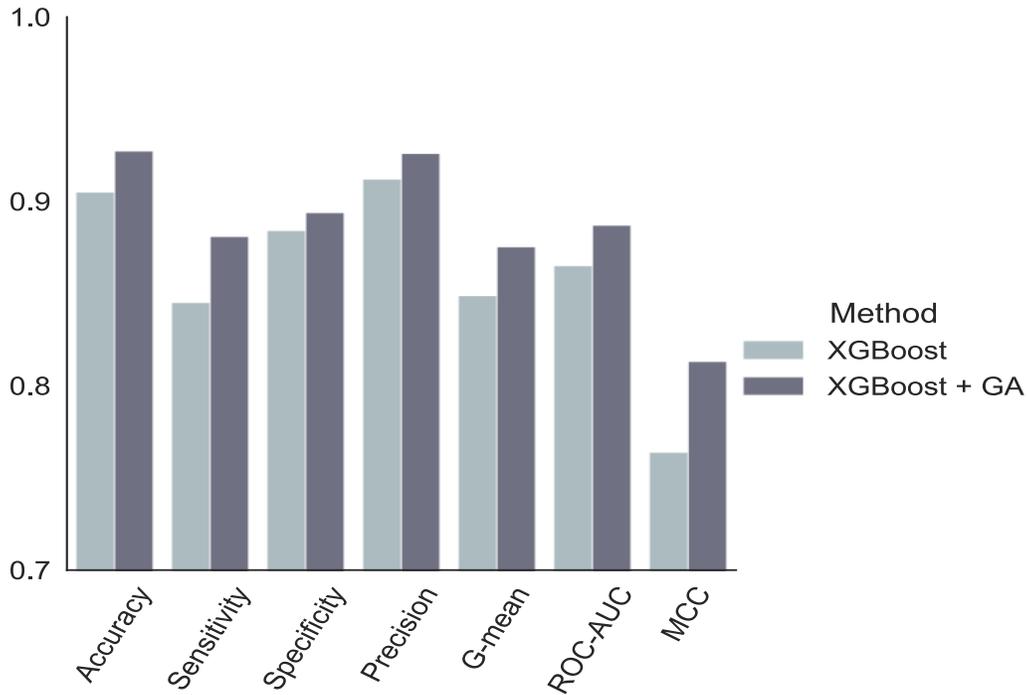

**Figure 4:** Improvement in performance after applying GA

**4.3 Performance comparison of the proposed approach with the other approaches presented in previous studies**

ROMA is a standard algorithm used in the prediction of risk in OC patients. It is based on serum biomarkers HE4 and CA125. We calculated the ROMA score for the premenopausal and postmenopausal patients using equations (3), (4), and (5) according to [11]. The performance metrics calculated from the obtained scores are presented in Table 5. Terlikowska et al. [29] proposed new cut-off values for the calculation of ROMA scores. They suggested 14.9% and 33.4% as the cut-off values for premenopausal and postmenopausal women, respectively. Using their proposed cut-off values, the ROMA score was calculated, and the measures obtained are reported in Table 5. Lu et al. [33] in their study proposed an ML-based prediction system using HE4 and CEA with logistic regression. We evaluated its performance, and the obtained results are provided in Table 5. Ahamad et al. [34] used the LGBM classifier for the prediction of OC. The performance measures obtained from the algorithm are presented in Table 5.

A comparison of the performance with previously reported studies as well as the ROMA algorithm is illustrated in Figure 5. As can be observed, our proposed approach outperformed the other techniques by a large margin. The average MCC score obtained in our study was 0.813 which is significantly higher than any of the other techniques. The overall accuracy and g-mean score of the model are 92.7% and 87.5%, respectively, which is also the highest among the studies. The ROMA algorithm is the current standard in predicting risk in OC patients. Although it performed well in the postmenopausal population, the algorithm provided quite a poor sensitivity score of only 56% for the premenopausal population. Sensitivity, also known as True Positive Rate (TPR), is a measure of goodness of the algorithm in accurately predicting the positive cases which is extremely important in medical diagnosis. Failing to identify the patients at risk of cancer can have fatal outcomes. The low sensitivity, as well as the low MCC score obtained from the ROMA algorithm, manifests the critical need to revisit the existing strategy and look for more effective biomarkers for accurate diagnosis. Terlikowska et al. [29] suggested a slight modification to the ROMA algorithm. Using the new cut-off values proposed by the authors, a small improvement in performance was observed. Lu et al. [33] first introduced ML in the diagnosis of OC. They suggested the use of a new biomarker CEA instead of CA125. Using HE4 and CEA, an overall improvement in performance was noticeable in the premenopausal population. However, the algorithm fails utterly in postmenopausal cases. The model fails to differentiate between the positive and negative cases and predicts all the cases as positive, resulting in a specificity score of 0. This indicates the need for separate biomarkers for the premenopausal and postmenopausal populations, which is not explored in any of the previous studies.

## 5  DISCUSSION

Of the 48 features in the original dataset, only a few of them have been identified by the GA as the most critical risk factors. It can also be observed that the risk factors vary with the menopausal status. To understand the overall contribution of those features in predicting OC, XAI techniques have been incorporated into this study. Figure 6 shows the significance of those features in the prediction of malignancy based on Shapley values. As can be observed from the figure, CEA, HE4, and ALB are found to be the most important predictors for the premenopausal population while CA125 is the most important for the postmenopausal population. Other features have contributions of different degrees in the overall prediction. The XAI techniques can further enable us to understand the quantitative relation of these features with the target variable and their individual effect in a particular prediction. Figure 7 demonstrates the positive or negative relationships between these risk factors with the outcome. A positive SHAP value indicates that the feature is increasing the risk of cancer, while a negative SHAP value is an indication of the reduction of risk. As can be observed from Figure 7 (a), a high value of HE4, CEA, and TP generally increases the risk of cancer in premenopausal patients, while a high value of ALB reduces the risk.

Table 5: Performance measures from different approaches proposed in previous studies.

| Approach | Menopause Status | Accuracy | Sensitivity | Specificity | Precision | G-mean | MCC |
|---|---|---|---|---|---|---|---|
| ROMA [11] | Premenopausal | 0.804 | 0.560 | 0.922 | 0.777 | 0.718 | 0.533 |
| | Postmenopausal | 0.916 | 0.906 | 0.956 | 0.988 | 0.931 | 0.776 |
| | Total Average | 0.860 | 0.733 | 0.939 | 0.883 | 0.824 | 0.654 |
| Terlikowska et al. [29] | Premenopausal | 0.813 | 0.52 | 0.955 | 0.848 | 0.705 | 0.556 |
| | Postmenopausal | 0.891 | 0.875 | 0.956 | 0.988 | 0.914 | 0.726 |
| | Total Average | 0.852 | 0.697 | 0.956 | 0.918 | 0.809 | 0.641 |
| Lu et al. [33] | Premenopausal | 0.826 | 0.483 | 0.993 | 0.987 | 0.672 | 0.603 |
| | Postmenopausal | 0.806 | 1 | 0 | 0.806 | 0 | 0 |
| | Total Average | 0.816 | 0.741 | 0.496 | 0.897 | 0.336 | 0.301 |
| Ahamad et al. [34] | Premenopausal | 0.852 | 0.710 | 0.921 | 0.845 | 0.802 | 0.669 |
| | Postmenopausal | 0.907 | 0.947 | 0.767 | 0.938 | 0.840 | 0.732 |
| | Total Average | 0.879 | 0.8285 | 0.844 | 0.891 | 0.821 | 0.7 |

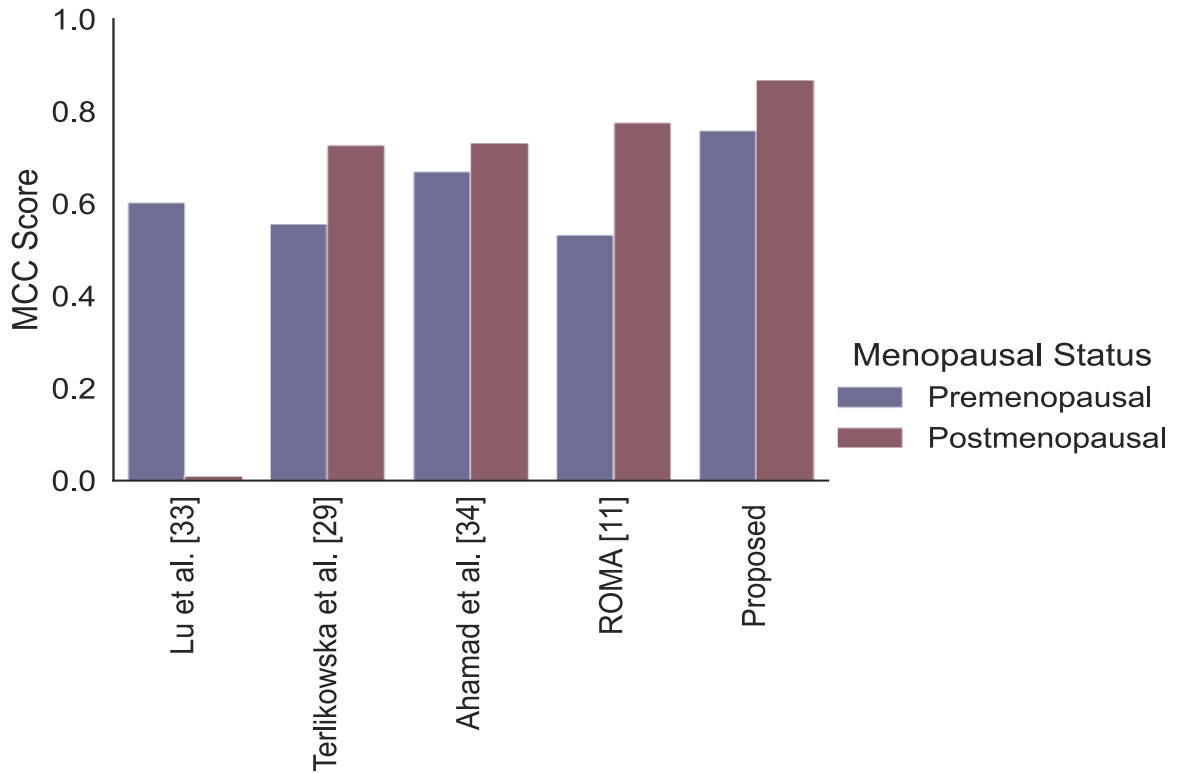

**Figure 5:** Performance comparison of the proposed approach with other approaches

### 5.1 Analysis of the risk factors identified by the GA in the diagnosis of OC

Only CA125 has been found as a common predictor in both groups which concurs with the existing diagnostic tools for OC [11], [12]. CA125 is considered to be the most promising biomarker for the screening of OC [45]. It is a type of tumor marker that measures the amount of a particular protein called Cancer Antigen 125 in the blood sample. A high level of CA125 is often found in the blood of OC patients. This is also apparent from Figure 7, where a high level of CA125 is found to have a positive Shapley value which indicates a risk of cancer.

HE4 is another biomarker that is commonly used for the risk assessment of OC in women. The menopausal status greatly influences the value of HE4 [46]. In our study, HE4 was found to be a suitable biomarker only for premenopausal patients. This agrees with the study conducted by Wang et al. [28] who suggested that HE4 is more useful in diagnosing OC in the premenopausal population. In a separate cohort study conducted on 1590 OC patients in the UK, Aleksandra et al. [47] evaluated the diagnostic value of HE4 in the prediction of OC and concluded that HE4 adds little value to the differential diagnosis of adnexal masses in postmenopausal women. These studies confirm the credibility of the findings from our study.

These two risk factors are jointly used in the ROMA algorithm for the diagnosis of cancer. Although it is a popular way of risk assessment in OC patients, it offers comparatively low sensitivity and precision, especially in premenopausal patients (Table 5). On account of that, there has been growing research for the identification of new biomarkers that can

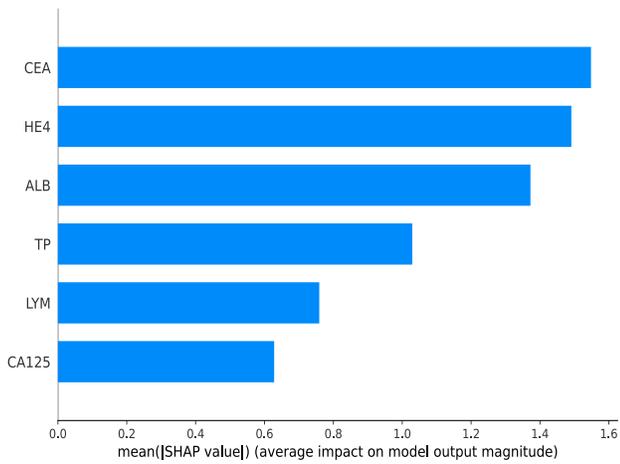
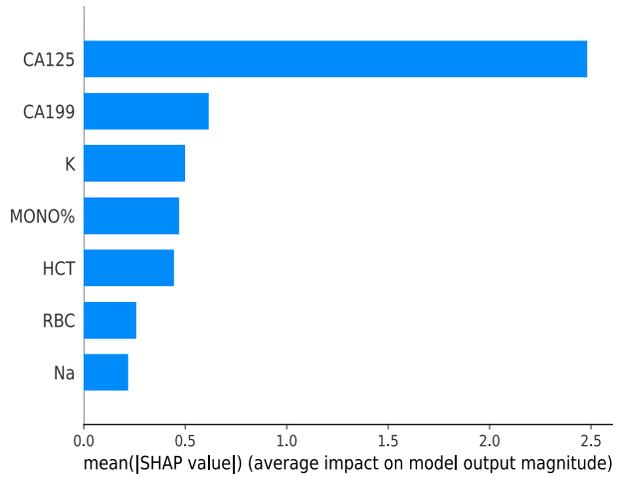

**(a)** **Premenopausal population**  **(b)** **Postmenopausal population**

**Figure 6**: Contribution of the selected features in the prediction of OC

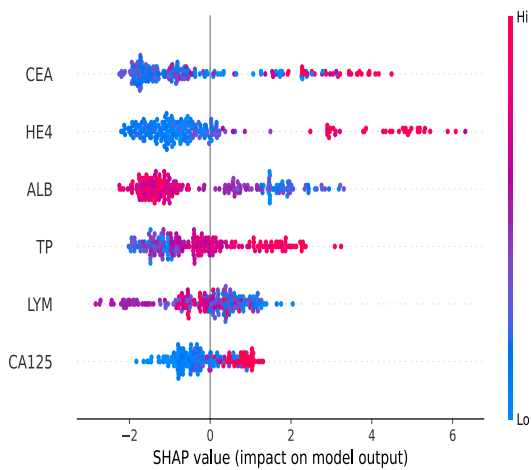
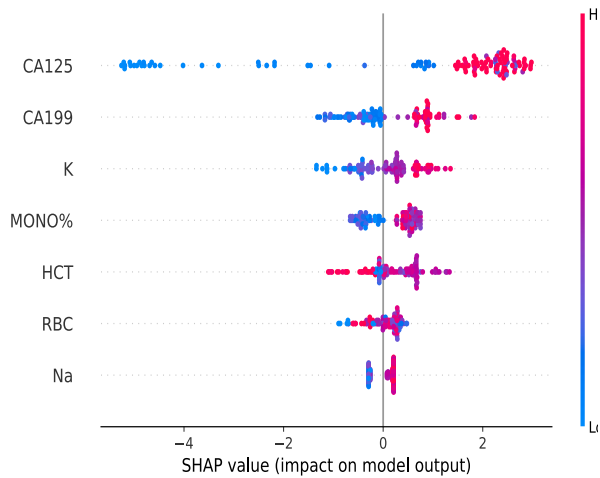

**(a)** **Premenopausal population**  **(b)** **Postmenopausal population**

**Figure 7**: Relationship of the predictor variables with the target

aid in the early diagnosis of OC with improved performance. Several independent studies can be found in the literature proposing new potential biomarkers for OC. However, these studies do not provide a unified framework for the diagnosis of OC but rather investigate the viability of a particular biomarker in the prediction of OC. From the experiments conducted in this study, we identify several new biomarkers as potential risk factors for the differential diagnosis of OC.

A low level of serum albumin (ALB) is found to be contributing positively to increasing the risk in patients (Figure 7). This can be associated with Hypoalbuminemia in cancer patients which results from malnutrition, low appetite, weight loss, and cachexia due to the host responses to the tumors. Hypoalbuminemia occurs when the body doesn't produce enough of the albumin protein. In an independent study conducted by Ge et al. [48], the authors also suggested that preoperative serum albumin can be used as an independent predictor of survival in epithelial OC patients.

For the postmenopausal population, cancer antigen CA19-9 is found to be the second most important predictor after CA125. An elevation of serum CA19-9 in the blood increases the risk of cancer (Figure 7 (b)). Similar findings have been reported in a retrospective study conducted by Lertkhachonsuk et al. [49]. The authors suggested that an elevation of the serum CA19-9, CA-125, CEA, and tumor size are useful predictors of OC. CEA is also found to be a critical predictor in our study for premenopausal patients. A high value of CA19-9 (>39 U/mL) or CEA (>3.8 U/mL) is strongly associated with OC malignancy.

Among other predictors, Monocyte ratio (MONO%), Hematocrit level (HCT), and Potassium (K) are found to be important biomarkers in postmenopausal women by GA. Total Protein (TP) and Lymphocyte count (LYM) are found as important biomarkers in premenopausal women by GA. These risk factors can also be linked to the diagnosis and prognosis of OC from different studies [50]–[53]. The contributions of some of the other attributes like Na or RBC are found to be quite limited by SHAP. The selection of these features can be attributed to the stochastic nature of the ML or FS techniques. By increasing the number of iterations or changing the parameter values of GA, small variations in the selected feature set can be observed. However, the most significant predictors remain unchanged. Thus, the inclusion of SHAP to assess and identify the risk factors from the pool of the most representative feature set provided by the GA can be considered a more effective and reliable way for the accurate diagnosis of OC.

### 5.2 Interpretation of the model using XAI (SHAP)

XAI provides transparency to the model prediction while also enabling result tracking. It helps in analyzing the model results and the radix of the predictions. SHAP is one of the most useful XAI tools and has been utilized in this study.

The variable importance plots provided in Figures 6 and 7 demonstrate the contribution of the features in predicting OC and how they are related (positively or negatively) to the target variable. To further analyze the relationships between the biomarkers and the risk of cancer and determine the threshold points, partial dependence plots can be utilized. Figures 8 and 9 show the partial dependence plots for some of the key risk factors in premenopausal and postmenopausal women, respectively. Here, Shapley values are utilized to quantify the relationships between the changes in risk with the changes in feature values.

These partial dependence plots demonstrate linear, monotonic, or more complex relationships between the features and the target. For instance, the risk of cancer can be found to be almost linearly increasing with an increase in the TP values (Figure 8). These plots automatically include another variable that interacts the most with the chosen variable. For instance, both ALB and CEA can be found to be correlated with HE4 (Figure 8). The abrupt changes in the risk (in terms of Shapley values) found in the figures can also be used to identify the trigger points. For instance, in the postmenopausal population, CA125 has negative Shapley values with a serum level < 30 U/mL. However, a sudden change in Shapley values (positive) can be observed after that point (Figure 9) which indicates an increase in risk. Similar thresholds can also be observed in other risk factors as well. However, the trigger points obtained here are not absolute. They are data-specific and can vary slightly on a different database.

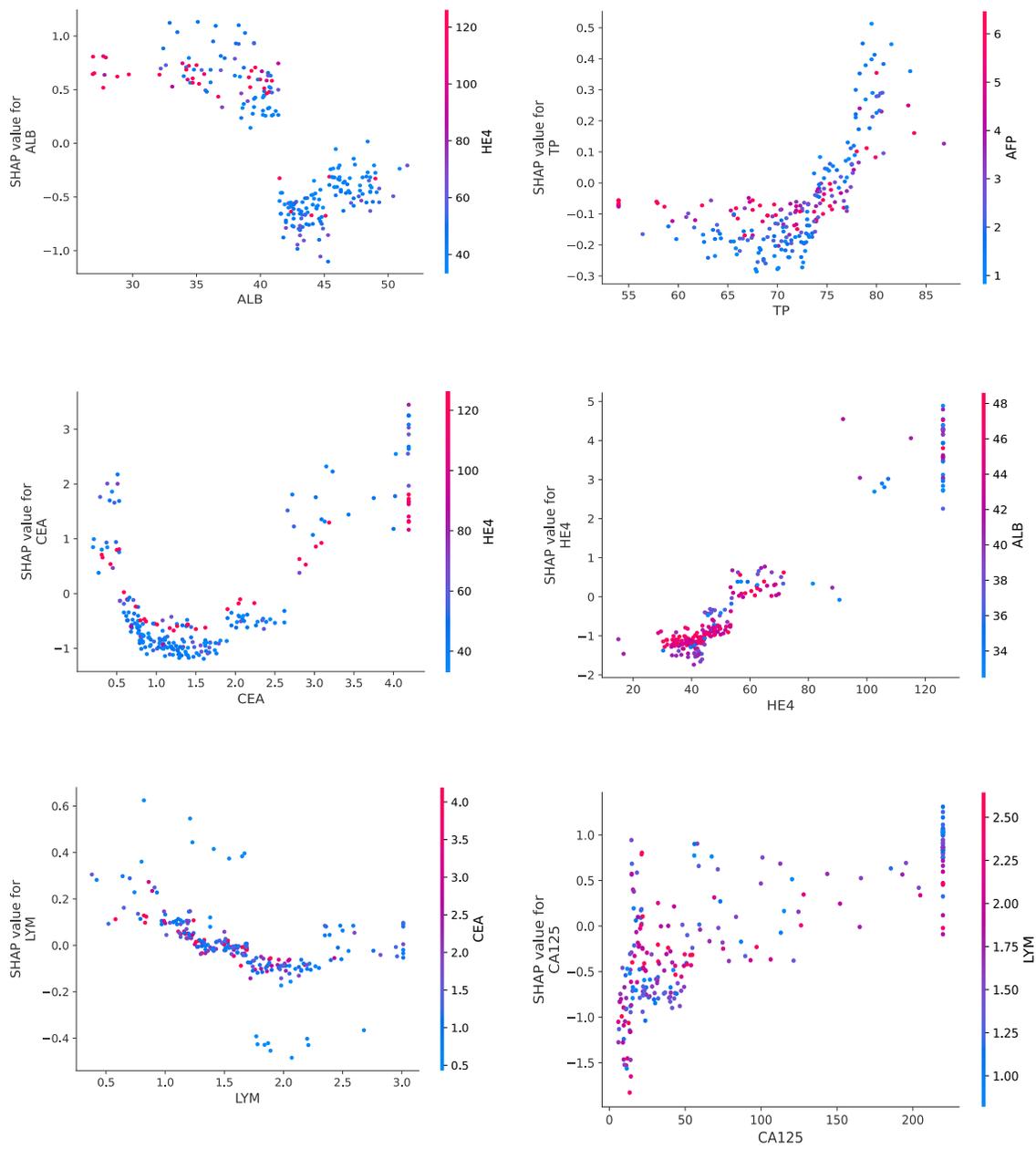

**Figure 8:** Partial dependence plots for risk factors in premenopausal patients

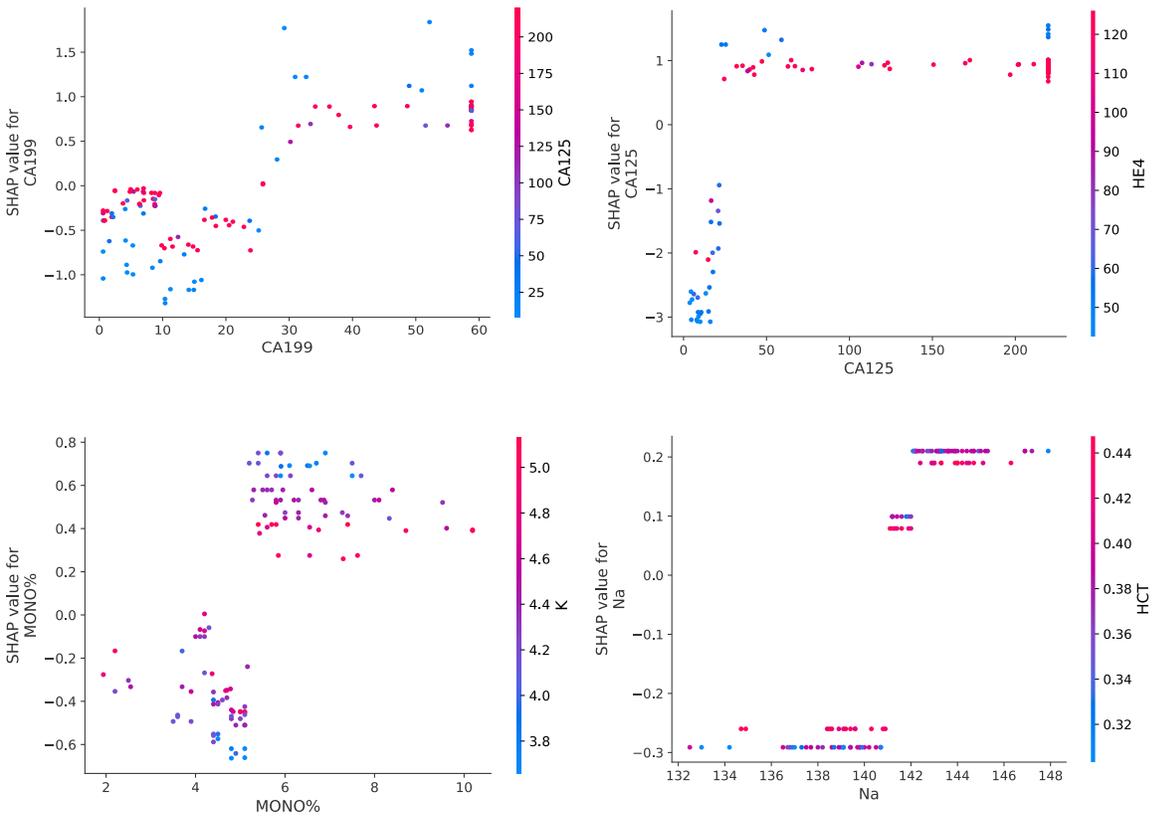

**Figure 9:** Partial dependence plots for risk factors in postmenopausal patients

### 5.3 Local interpretability using SHAP

SHAP can be utilized to interpret individual predictions made by the ML classifier and provide clinicians with insights into the factors leading to that prediction. This helps clinicians evaluate the prediction made by the model for a patient and provide a more informed decision rather than blindly trusting the results of the model. Figure 10 provides the interpretation of model prediction results for two patients.

Figure 10 (a) interprets the prediction result of a premenopausal woman with a high risk of cancer. The individual effect of the risk factors in pushing the prediction from a base value of 0.26 to 0.9 can be observed from the figure. HE4 has the highest contribution followed by ALB and CA125 in pushing the prediction rightwards, while the other three variables can be seen contributing negatively in formulating the final prediction. The patient has a high value of HE4 and a low value of ALB which are important predictors of OC as can be observed in Figure 8. Similarly, Figure 10 (b) interprets the prediction result of a postmenopausal woman with a low risk of cancer. CA125 can be seen as the prominent factor leading the prediction toward zero probability. A low value of CA125 has been identified as an important factor in minimizing the risk of cancer (Figure 9).

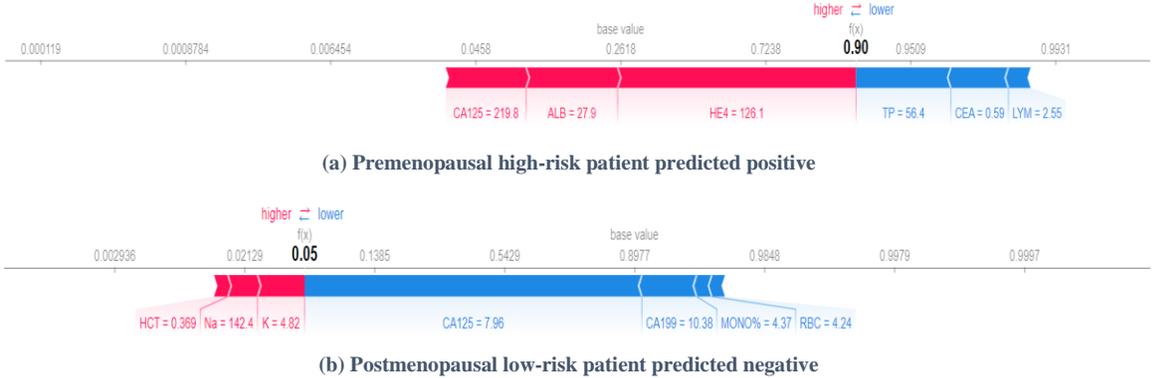

(a) Premenopausal high-risk patient predicted positive

(b) Postmenopausal low-risk patient predicted negative

**Figure 10:** SHAP force plot for the interpretation of individual predictions

SHAP force plots can be further utilized to interrogate the wrong predictions made by the model and understand the reasons behind them. Two such false-positive predictions are shown in Figure 11. Figure 11 (a) shows the result of a low-risk premenopausal patient which the model predicted positive with a probability of 0.58. A value so close to 0.5 can already raise suspicion of the clinician. The patient has an HE4 value of 46.27, a CEA value of 1.03, and a TP value of 71.3, all of which are indications of low risk and thus pushing the prediction leftward. However, the rightward contributions of the other three variables push the decision boundary toward positive. The ALB level of 41.4 is very close to the cut-off point which makes it difficult for the model to correctly assess. On the other hand, the LYM ratio of 1.09 is an unusual point in Figure 8 where some data points do not follow the usual trend. This can be due to the natural variation in the data or some mistakes in recording the data point which necessitates further manual investigation. Thus, SHAP force plots provide a way to investigate the predictions made by the model and aid clinicians in critical decision-making.

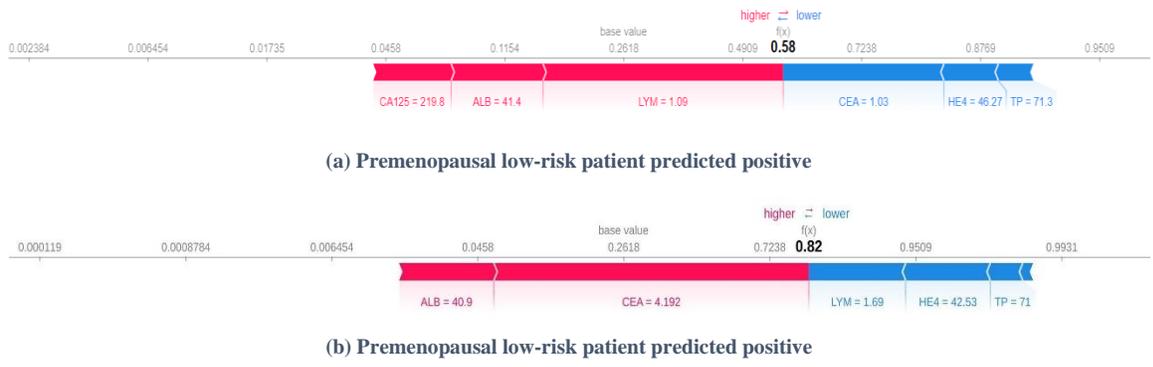

(a) Premenopausal low-risk patient predicted positive

(b) Premenopausal low-risk patient predicted positive

**Figure 11:** Interpretation of wrong predictions made by the model using SHAP force plot

## 6 CONCLUSION

Ovarian cancer is one of the most prevalent types of cancer in women with a high mortality rate. Early and accurate diagnosis is crucial for the survival of the patients. However, it often goes undiagnosed until it has progressed to advanced stages due to the lack of more appropriate and effective biomarkers. The early screening tools lack proficiency with a high misclassification rate. This calls for a systemic study to identify new potential biomarkers and develop a more robust screening tool that can predict the existence of cancer with higher accuracy. AI techniques have been increasingly applied in the field of health informatics to aid clinicians in providing better healthcare. In this study, we developed an ML-based diagnostic system for the reliable detection of OC. Explainable AI tools have been incorporated to improve the efficacy of the decision-support framework for clinicians. New biomarkers have been identified using a combination of GA and SHAP for a more effective diagnosis of OC.

The study has been conducted on a dataset containing records of patient demographics, blood routine tests, tumor biomarkers, and general chemistry test results. As menopausal status has been identified as a critical factor affecting the symptoms and development of OC, experiments were conducted separately on the premenopausal and postmenopausal populations. GA has been employed to identify the most representative feature set from the data. Unlike previous studies which sought a common biomarker, we identified different risk factors for premenopausal and postmenopausal women. CA125 is a common predictor in both groups, however, biomarkers like HE4 should only be considered in the premenopausal population while CA199 should be considered for the postmenopausal population only.

An advanced ensemble ML algorithm, XGBoost classifier, was trained on the selected features and the performance was evaluated using a 10-fold cross-validation technique. ROC-AUC scores of 0.864 and 0.911 were obtained on the premenopausal and postmenopausal populations, respectively. The diagnostic accuracy obtained from the proposed approach outperforms the existing methods, along with the standard ROMA method, by a significantly large margin. To provide transparency to the model predictions, a popular XAI tool SHAP was utilized to interpret the model predictions and understand the individual contribution of the selected features behind the decisions provided by the ML model. With the help of SHAP visualization techniques, the most critical risk factors, and their trigger points are determined. CEA, HE4, and ALB are identified as the most important biomarkers in premenopausal women while CA125 and CA19-9 are identified as the most important biomarkers in postmenopausal women. These findings can provide a new perspective in the search for novel predictors for the effective diagnosis of OC. However, it requires further clinical research on a larger population from diverse geographical origins. It could not be conducted due to the lack of such available data. Since the process is entirely data-driven, any inherent bias present in the data would be reflected in the results. Therefore, we aim to collect more data in the future and conduct further studies to eliminate any geographical bias.

The superior result achieved by our proposed framework signifies the system's ability to provide more accurate predictions. The established hybrid decision-support framework with the incorporation of SHAP enhances the reliability of the system, allowing clinicians to make more informed decisions. The proposed framework realizes the potential of AI by providing a system that can be considered safe, transparent, and beneficial in the early diagnosis of OC.